\documentclass[runningheads]{llncs}

\usepackage[dvipsnames]{xcolor}
\usepackage{amsmath, amssymb}
\usepackage{graphicx}
\usepackage{xspace}
\usepackage{svg}
\usepackage{algorithm}
\usepackage{algpseudocode}
\usepackage[hidelinks]{hyperref}
\usepackage[acronym]{glossaries}
\usepackage{subcaption}
\usepackage{array}
\usepackage{geometry}
\usepackage[style=base]{caption} 
\usepackage{cite}
\usepackage[export]{adjustbox}
\usepackage{multirow}

\newcommand{\ie}{{i.e.}\xspace}

\newcommand{\wrt}{{w.r.t.}\xspace}
\newcommand{\eg}{{e.g.}\xspace}

\makeglossaries
\newacronym{cf}{CF}{Catastrophic Forgetting}
\newacronym{lrp}{LRP}{Layer-wise Relevance Propagation}
\newacronym{task-il}{Task-IL}{Task-incremental learning}
\newacronym{xai}{XAI}{Explainable Artificial Intelligence}
\newacronym{rnf}{RNF}{Relevance-based Neural Freezing}

\usepackage{enumitem,amssymb}
\newlist{todolist}{itemize}{2}
\setlist[todolist]{label=$\square$}
\usepackage{pifont}

\usepackage[normalem]{ulem}

\begin{document}

\title{Explain to Not Forget: Defending Against Catastrophic Forgetting with XAI}

\author{Sami Ede\inst{1}$^{,\dag}$ \and Serop Baghdadlian\inst{1}$^{,\dag}$ \and Leander Weber\inst{1} \and
An Nguyen\inst{2} \and Dario Zanca\inst{2} \and \\
Wojciech Samek\inst{1,3,4}$^{,\star}$ \and Sebastian Lapuschkin\inst{1}$^{,\star}$}
\authorrunning{S. Ede et al.}

\institute{$^1$ Fraunhofer Heinrich Hertz Institute, 10587 Berlin, Germany \\
$^2$ Friedrich Alexander-Universität Erlangen-Nürnberg, 91052 Erlangen, Germany\\
$^3$ Technische Universit\"at Berlin, 10587 Berlin, Germany \\
$^4$ BIFOLD – Berlin Institute for the Foundations of Learning and Data, 10587 Berlin, Germany \\
$^\dag$ contributed equally \\
$^\star$ corresponding authors: \email{\{{sebastian.lapuschkin, wojciech.samek\}}@hhi.fraunhofer.de}
}

\newcommand{\sami}[1]{#1}
\newcommand{\slap}[1]{#1}
\newcommand{\sweb}[1]{#1}
\newcommand{\ft}[1]{#1}

\maketitle

\begin{abstract}

The ability to continuously process and retain new information like we do naturally as humans is a feat that is highly sought after when training neural networks.
Unfortunately, the traditional optimization algorithms often require large amounts of data available during training time and updates \wrt new data are difficult after the training process has been completed.
In fact, when new data or tasks arise, previous progress may be lost as neural networks are prone to catastrophic forgetting.
Catastrophic forgetting describes the phenomenon when a neural network completely forgets previous knowledge when given new information.
We propose a novel training algorithm called \textit{Relevance-based Neural Freezing} in which we leverage Layer-wise Relevance Propagation in order to retain the information a neural network has already learned in previous tasks when training on new data.
The method is evaluated on a range of benchmark datasets as well as more complex data.
Our method not only successfully retains the knowledge of old tasks within the neural networks but does so more resource-efficiently than other state-of-the-art solutions.


\end{abstract}

\section{Introduction}
While neural networks achieve extraordinary results in a wide range of applications,
from the medical field to computer vision or successfully beating humans on a variety of games \cite{Wilm2022, silver2016alpha},
the established training process typically relies on a large amount of data that is present at training time to learn a specific task. For example, the famous ImageNet dataset \cite{deng2009imagenet} consists of more than 14 million images which results in a size of more than 150 GB, while the authors of \cite{Radford2021Clip} collected a dataset of 400 million images that make up more than 10 TB of data \cite{Schuhmann2021LAION}. 
Large amounts of samples can help models generalize better by avoiding overfitting in single examples, but in turn make model training extremely expensive.
If more data is later added, and the model should be able to correctly predict on both new and old data,
usually it has to be finetuned or trained from scratch with the expanded dataset as opposed to only the new data. Otherwise, \textit{catastrophic forgetting} \cite{French1999CF} can occur when learning multiple consecutive tasks or from non-stationary data.
One prominent example of this is reinforcement learning,
in which an agent continuously interacts with its environment,
using a stream of observations as training data.
As the observations change with the agent's actions in the environment, the data distribution becomes non-i.i.d., leading to catastrophic forgetting in the agent that is usually countered with an ``experience buffer'',
in which earlier observations are saved. These saved observations are then randomly repeated during training.
Other applications would also benefit from solutions to continuous or lifelong learning,
e.g., 
medical applications such as skin cancer detection, where more targets could be added after additional samples have been obtained.
The issue of catastrophic forgetting is especially pronounced when previous data is not accessible anymore, \eg, due to being proprietary, making retraining impossible.

Recently, techniques of \gls{xai} \cite{SamPIEEE21} have been proposed which are able to identify the elements of a neural network model crucial for solving the problem \sweb{a} model has been optimized for.
One such method is \gls{lrp} \cite{bach2015LRP}, which assigns relevance scores to latent network structures through modified backpropagation. In the recent past, this information has been used with great success to efficiently and effectively reduce neural network complexity without sacrificing performance \cite{yeom2021pruning,BecXXAI22}.

In this paper, we are proposing \textit{\gls{rnf}, }a novel approach to alleviate catastrophic forgetting that builds upon the aforementioned pruning technique.
Instead of compressing the network, the information about unit importance is used to freeze the knowledge represented by the learned network parameters by inhibiting or completely stopping the training process for those parts of the \slap{model} that are relevant for a specific task, while the remaining units are free to learn further tasks. 
We evaluate our method on several commonly used datasets, \ie, MNIST \cite{deng2012mnist}, CIFAR10 and CIFAR100 \cite{krizhevsky2009cifar}, ImageNet \cite{deng2009imagenet}, and the challenging Adience dataset for facial categorization \cite{EidingerAdience2014}, which is a dataset of photos shot under real-world conditions, meaning different variations in pose, lighting conditions and image quality.

\section{Related Work}

In this section we briefly review the theoretical background of explainable AI and catastrophic forgetting.

\subsection{Explainable Artificial Intelligence}
In recent years, \gls{xai} has gotten more and more attention as the urgency to understand how ``black box'' neural networks arrive at their predictions has become more apparent. Especially in applications that have far-reaching consequences for humans, like the prediction of cancer (\eg~\sami{\cite{Chereda2020GNNLRP, Haegele2020histopathological, EVANS2022281}}), it is not only important to know what the network predicted, but also why \slap{a certain} decision \slap{has been} made.
Generally, methods from \gls{xai} can be roughly divided into two categories: %

\textit{Global} explanations provide general knowledge about the model, its feature sensitivities and concept encodings.
Some approaches aim to identify the importance of specific features, concepts\sweb{,} or data transformations (\eg~\cite{guyon2002,10.5555/944919.944968,Kim2018InterpretabilityBF}) by analyzing the model's reaction to real or synthetic data, while others try to assess their model by finding important neurons and their interactions \cite{Hohman2019summit}, or \sweb{by} finding the concepts encoded by hidden filters through synthesizing the input that maximizes their activation \cite{Bengio2009visualizing, nguyen2016synthesizing, olah2017feature}. 

Instead of providing insight \sweb{into} the model\slap{'}s general understanding of the data, \textit{local} explanation methods aim at making individual model predictions interpretable, \ie, by ranking the importance of features \sweb{\wrt} specific samples.
By attributing importance scores to the input variables, these explanations can be illustrated as heatmaps with the same dimensions as the input space.
Among the local explanation methods, there are again multiple approaches, some still treating the model as a ``black box'', approximating the local explanations via separately trained proxy models \cite{ribeiro2016why, Zeiler2013Visualizing} or otherwise applying perturbation or occlusion techniques \cite{Fong2017meaningfulpert,  zintgraf2017visualizing}.
Other methods use (augmented) backpropagation in order to compute the importance ranking of the input or latent features, such as \cite{bach2015LRP, baehrens2009explain, 2017DTD, sundararajan2017axiomatic}.
Our proposed method leverages the advantages of Layer-wise Relevance Propagation~\cite{bach2015LRP}\sweb{,} as \sweb{this} method's ability to measure the per-prediction usefulness and involvement of \slap{(also latent)} network elements has recently shown great success \cite{yeom2021pruning,BecXXAI22} \slap{in applications for model improvement}.

\subsection{Catastrophic Forgetting}
Unlike humans or animals, neural networks do not have the inherent ability to retain previously attained knowledge when they are presented with new information while being optimized.
This effect is characterized by a drastic performance decrease on tasks trained earlier when progressing on a new task or dataset.
This phenomenon is described by the term catastrophic forgetting
\cite{French1999CF}.
As neural networks are generally assumed to train on i.i.d. data, adding a new task to be learned can violate this assumption, causing the gradient updates to override the weights that have been learned for the previous tasks and causing the aforementioned loss of old knowledge. One way to combat catastrophic forgetting is experience replay \cite{delange2019contlearning}, in which the old data is interspersed with the new data, simulating i.i.d. data such that the network retains the old knowledge. However, this approach is inefficient, does not allow online-learning\sweb{,} and may even be impossible if access to the old data is not available. Therefore, numerous approaches have been proposed to tackle this problem more efficiently.
 The approach of \cite{wortsman2020supermasks} learns masks defining subnetworks in untrained models that are responsible for a given task\sweb{,} while \cite{serra2017hardattention} concurrently learn binary attention vectors to retain the knowledge obtained in previous tasks.
Other approaches \ft{(\eg}, \cite{kirkpatrick2017overcoming,Zenke2017synapticintelligence}\ft{)} propose constraints on weight updates for neurons that have been identified as being pertinent for a previous task.
Dynamically Expandable Networks \cite{Lee2017expandablenetworks} increase the network capacity when training a new task.

In this paper, we propose a training algorithm that ---  motivated by the successful \gls{xai}-based pruning method described in \cite{yeom2021pruning} --- uses \gls{lrp} in order to identify those neurons that are relevant for a given task.
After finding the important neurons, they are given a lower elasticity for learning subsequent tasks, such that the network efficiently retains the knowledge from previous tasks while still being able to learn additional tasks.

\section{\sami{Relevance-based Neural Freezing}}

 As a local attribution method, \gls{lrp} has shown \sami{(see} \cite{SamekBMBM15, yeom2021pruning}\sami{)} to not only deliver accurate and interpretable explanations about the input variables, the conservatory property of the local distribution rules also allows to gain insights on the importance of individual latent neurons and their associated filters.
 Additionally, \gls{lrp} is scalable \wrt~network depth, easy to implement through existing frameworks (\eg,\cite{anders2021software,Alber2019innvestigate}), and efficient with a linear computational cost \wrt~a backpropagation pass. It works by treating the prediction of the model $f(\textbf{x})$ \wrt~a network output of interest as the total sum of importance, or \textit{relevance} $R$, that is then redistributed towards the input variables:
 After the forward pass, the layers of a classifier are reversely iterated, redistributing the relevance among its neurons proportionally to their contributions in the preceding forward pass.
 This redistribution process follows a conservatory constraint, meaning that the sum of relevance in each layer is equal to the total amount of relevance at the model head:
 \begin{equation}
     f(\textbf{x}) = \dots = \sum_{d_i \in (l+1)}R_{d_i}^{(l+1)} = \sum_{d_j \in (l)}R_{d_j}^{(l)} = \dots = \sum_{d_k \in (l_0)}R_{d_k}^{(l_0)}~,
 \end{equation}
 where $f(\textbf{x})$ is \sweb{the} model output and $R_d^{l}$ is the relevance score of unit $d$ in layer $l$. 
Depending on the application and layer type, \sweb{various propagation rules with specific purposes} have been proposed.
For example, the \gls{lrp}-$\varepsilon$ rule \cite{samek2017understanding} is defined as 
\begin{equation}
    R_{j \leftarrow k} = \sum_k \frac{a_j w_{jk}}{\sum_{0,j} a_j w_{jk} + \varepsilon} R_k.
\end{equation} 
with $a_j$ being the input activation at neuron $j$, $w_{jk}$ being the learned weights between neurons $j$ in the lower- and $k$ in the upper layer and $\varepsilon$ being a small \slap{term of the same sign as the sum in the demoniator}.
This rule is typically used in upper layers to filter out weaker or contradictory values via the added $\varepsilon$ in the denominator, which results in smoother, less noisy heatmaps \slap{and prevents numerical instabilities}.
A discussion of other rules and their applications can be found in \cite{samek2017understanding}.
In this paper, we use the \gls{lrp}-$\varepsilon$ rule for fully-connected layers and the \gls{lrp}-$z^+$ rule for convolutional layers, as recommended in~\cite{kohlbrenner2020towards}.

\begin{figure}
    \centering
    \includegraphics[width=\columnwidth]{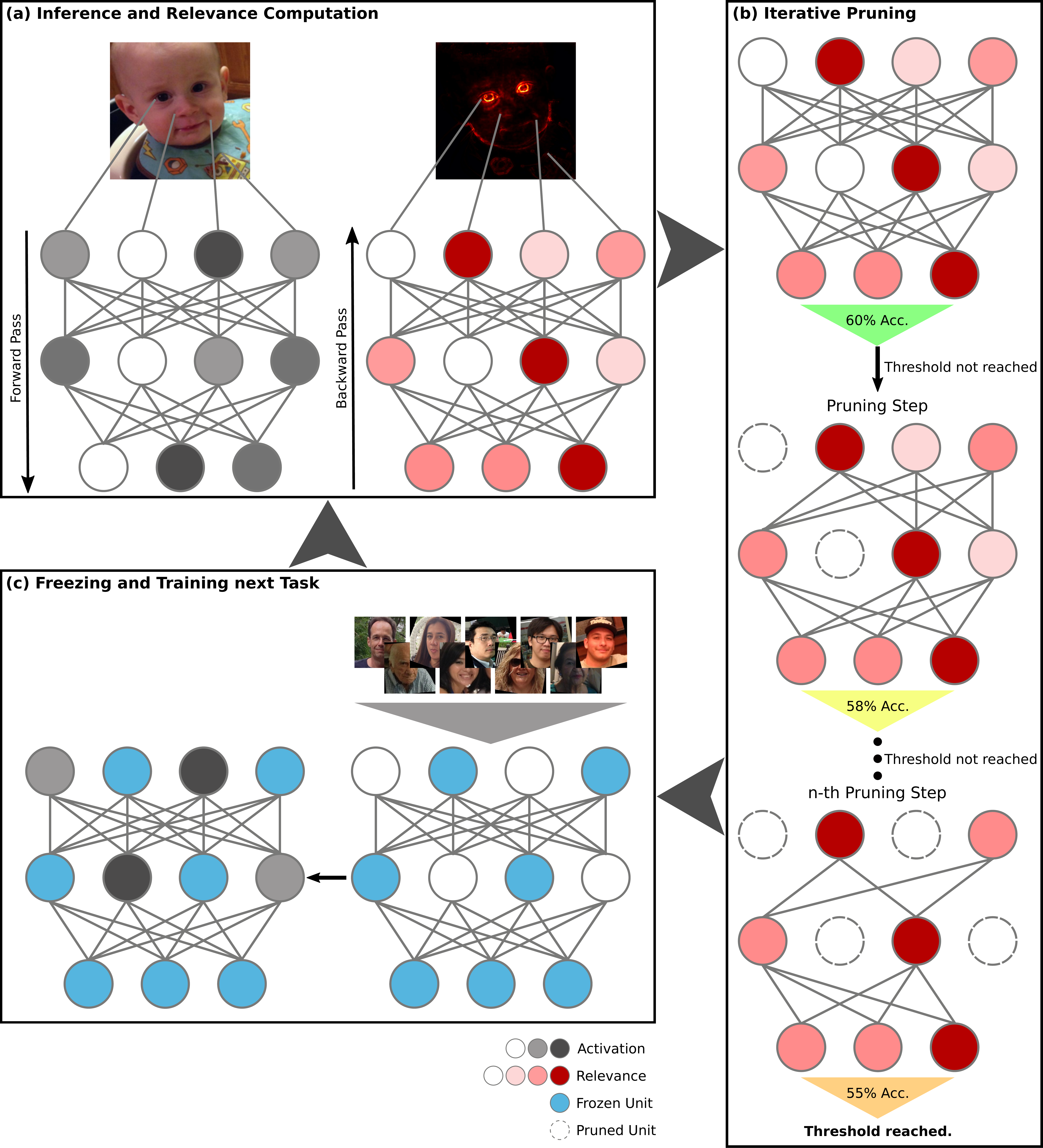}
    \caption{Illustration of our method to identify relevant neurons and freeze their training progress after training on a task. (a) Forward pass through the network trained on all previous tasks with a reference image of the current task (\emph{left}) and computation of \slap{network} unit importance using LRP (\emph{right}). (b) Iteratively set connections of irrelevant neurons/units to zero until performance threshold is reached. (c) Freezing the knowledge about the current task by setting the learning rate of the remaining units to zero and training on the next task. The process is repeated after training each task.}
    \label{fig:method}
\end{figure}

The proposed method aims to prevent catastrophic forgetting by decreasing the plasticity of neurons rated as important for a given, already optimized task. The general procedure is as follows: %
After training the model on the first task, the relevant units are identified by using \gls{lrp} on a small\sami{, randomly sampled} subset of the \sweb{(test-) }data, from here on called the \sami{\textbf{reference dataset}}, similarly as in \cite{yeom2021pruning}. \ft{Note, however, that in our application this reference data is selected specific to each task, containing only samples from classes that are learned in the current task. \gls{lrp}-attributions are computed \wrt the respective ground truth label of each sample.}
Until the model performance on the test set decreases by a \ft{predetermined} threshold, the units with the lowest relevance (computed on the reference dataset) are repeatedly selected and then pruned by setting their outgoing connections to zero.
Once the threshold is reached, the remaining units are assigned a lower learning rate for any subsequent tasks, as they were the most important \sweb{ones} for the current task.
To completely freeze the units, the learning rate is set to zero, but it is also possible to just lower the elasticity to a fraction of the original learning rate.
To continue training, the connections to the less relevant units are restored to their state before the pruning.
This procedure is outlined in Algorithm~\ref{alg:alg1}, \slap{and} an intuitive illustration can be found in Figure~\ref{fig:method}.

\begin{algorithm}[H]
\caption{\sami{Relevance-based Neural Freezing}}\label{alg:alg1}
\begin{algorithmic}

\Require untrained model \texttt{net}, reference data $\mathtt{x_r}$, task specific training data $\mathtt{x_t}$, pruning threshold \texttt{t}, pruning ratio \texttt{r}, task number $\mathtt{N_t}$, learning rate \texttt{lr}, learning rate for relevant units $\mathtt{lr_{frozen}}$, learning rate for irrelevant units $\mathtt{lr_{irrelevant}}$, epochs $\mathtt{N_e}$, network \texttt{units}, with \texttt{unit} $\in$ \slap{\{}\texttt{neurons}, \texttt{filters}\slap{\}}.
\For{\texttt{task} in $\mathtt{N_t}$}
    \For{\texttt{epoch} in $\mathtt{N_e}$}
        \State $\triangleright$ train \texttt{net} on $\mathtt{x_t}$ using \texttt{lr}
    \EndFor
    \For{\textbf{all} \texttt{layer} \textbf{in} \texttt{net}}
        \For{\textbf{all} \texttt{units} in \texttt{layer}}
            \State $\triangleright$ compute importance of \texttt{units} using \textit{LRP}
        \EndFor
    \EndFor
    \State $\triangleright$ sort \texttt{units} in descending order \wrt their global importance to the task
    \While{\texttt{t} not reached}
        \State $\triangleright$ zero out \texttt{r} \texttt{units} from \texttt{net} \textbf{where} \texttt{importance} is minimal
    \EndWhile
    \State \slap{$\triangleright$ mark remaining \texttt{units} in \texttt{net} as \texttt{relevant units}}
    \State $\triangleright$ lower elasticity of \slap{\texttt{relevant units}} for current task
    \State $\mathtt{lr_{\texttt{relevant units}}} \gets \mathtt{lr_{frozen}}$
    \State $\mathtt{lr_{\texttt{irrelevant units}}} \gets \mathtt{lr_{irrelevant}}$
    \State $\triangleright$ restore zeroed out connections to continue training.
\EndFor
\end{algorithmic}
\end{algorithm}

The stopping criterion for the unit selection can be determined freely and is not limited to network performance. For instance, it could also be the number of remaining free units in the network or even the energy consumption of the network when running inference. In the following experiments, the learning rate for all frozen units is set to zero.

\section{Experiments}
We show the effectiveness of the proposed technique on a number of increasingly difficult datasets and tasks. We start by using well-known toy and benchmark datasets, namely MNIST \cite{deng2012mnist}, as well as a combination of CIFAR10 and CIFAR100 \cite{krizhevsky2009cifar}, to illustrate the method and its conceptual functionality before also showing effectiveness in larger benchmark and real-world datasets.  We set the pruning threshold to 2\% \ft{accuracy lost relative to the accuracy after finetuning on the task}, which we found to be optimal during our experiments. \sami{The optimal value for this parameter can be determined using grid search. Using 2\%} results in enough free network capacity to learn the additional tasks while keeping the accuracy of the classifier as high as possible. Results on MNIST, CIFAR10, and CIFAR100 are averaged \sweb{over} 20 \slap{random} seeds, while results on the ImageNet and the Adience dataset are averaged over five \slap{random} seeds. In addition to the accuracy, we also evaluate the free \sami{\textbf{capacity}} of the model, \sami{which we define} as the percentage of unfrozen \sweb{\slap{network} units} after each task. Details on each experimental setup can be found in the Appendix \ref{appendix:expdet}. \sami{In all experiments, new tasks are introduced with a  \textit{task-incremental (\acrshort{task-il}}) setup \cite{Oren_2021_ICCV}. In a task-incremental setup, the neural network is additionally informed about the task identity during training as well as during inference. Each task has its own (separate) head, while the rest of the neurons are shared among all tasks.}

\subsection{MNIST-Split}
\label{sec:mnistsplit}
The first series of experiments is performed on the popular MNIST dataset. The dataset is split up into five tasks, each task being the classification of two digits, e.g., the first task consists of the digits 0 and 1, task two contains the digits 2 and 3, and so on.
The model is trained on the first task and then \sweb{finetuned} sequentially using a task-incremental setup, which we refer to as \sami{naive finetuning}. 

Figure \ref{fig:MNIST:split} shows the effect of \sami{\gls{rnf}} on the MNIST-Split dataset. The mean test accuracy over all tasks is increased by about 4\%, \slap{compared to naive finetuning} \ft{(Figure \ref{fig:MNIST:split_mean_acc})}, which is the most evident in the accuracy for both task one and task two. Instead of a drop of 30\% in accuracy, the model can still classify task one with an accuracy of almost 90\% and retains an accuracy for task two of over 90\%. \sami{The increase in baseline \slap{(naive fine-tuning)} accuracy after task five \slap{may} be attributed to a similarity
between the shapes of digits in tasks one and five: both the digits 8 and 9 have
rounded forms which makes it easier to distinguish between a 0 and a 1.} \sami{In the following experiments, all changes to the accuracy are reported in comparison to the naive finetuning baseline.}

\subsection{MNIST-Permuted}
The MNIST-Permuted setup increases the complexity of the MNIST dataset by introducing random pixel permutations to the digits. It is commonly used \sami{(see \cite{Farquhar2018robustevals})} in a ten-task configuration such that the unpermuted dataset poses as the first task while the remaining nine tasks \slap{consist of} different permutations of the original digits. The permutations are the same for all classes but change between tasks. 

\begin{figure}[!ht]
    \centering
    \subfloat[\centering Average Test Accuracy]{\includegraphics[width=0.5\columnwidth]{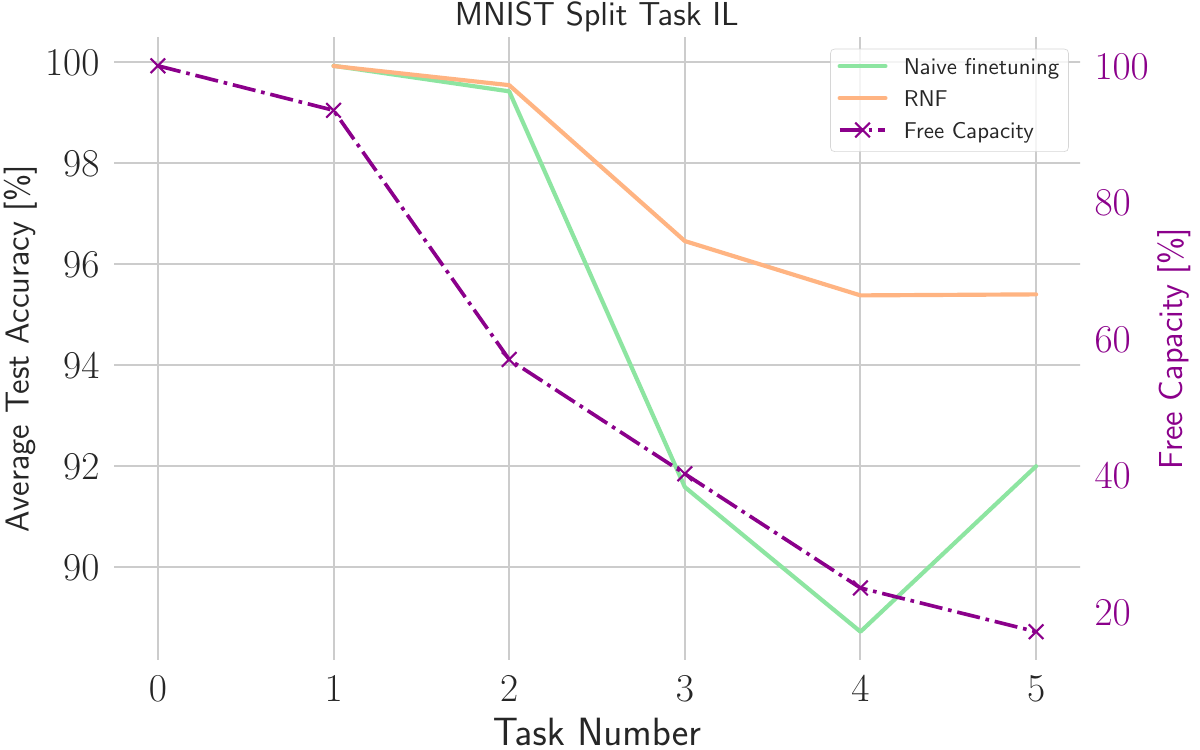}
    \label{fig:MNIST:split_mean_acc}} %
    \subfloat[\centering Mean test accuracy per task]{\includegraphics[width=0.5\columnwidth]{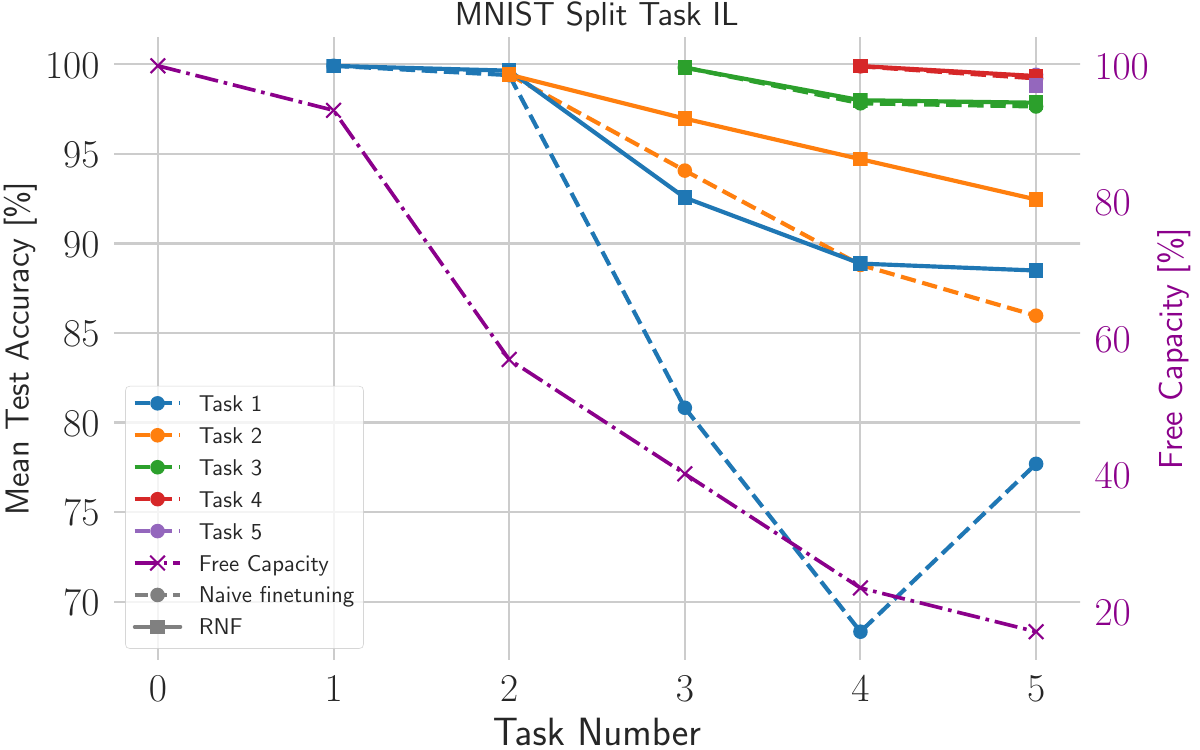}
    \label{fig:MNIST:split}} \\
    \subfloat[\centering Average Test Accuracy]{\includegraphics[width=0.5\columnwidth]{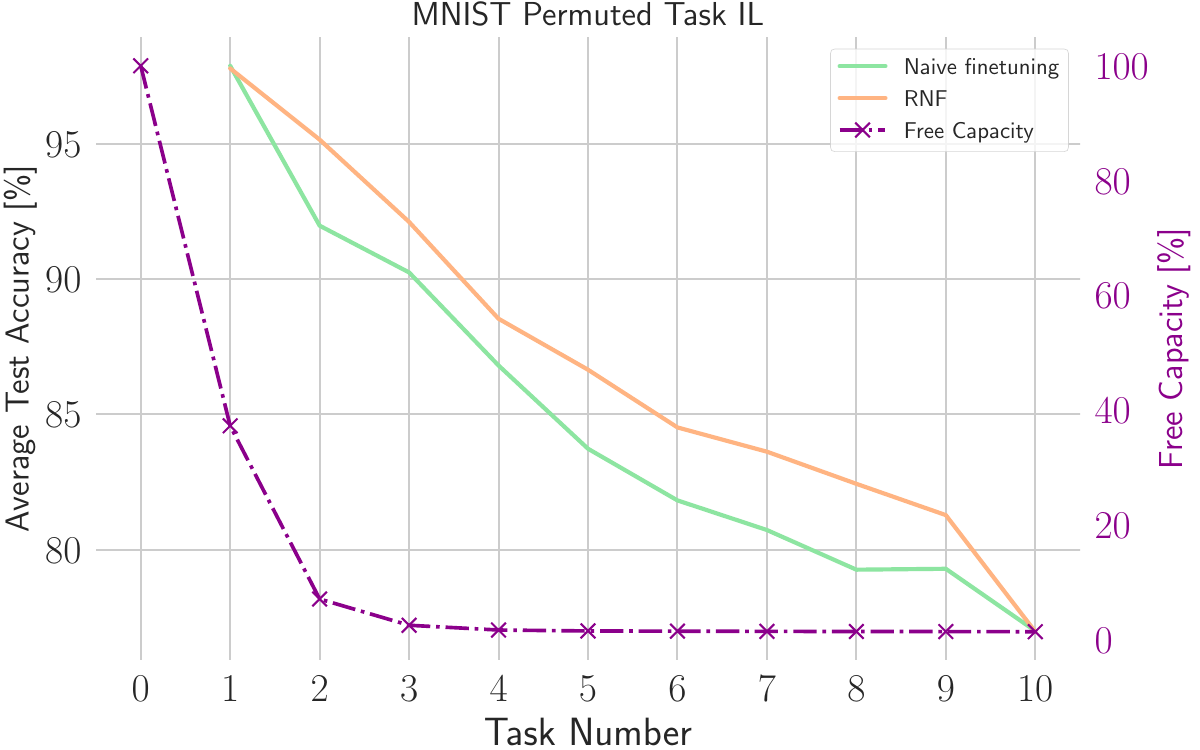}
    \label{fig:MNIST-Permuted:avg_acc}} %
    \subfloat[\centering Mean test accuracy per task]{\includegraphics[width=0.5\columnwidth]{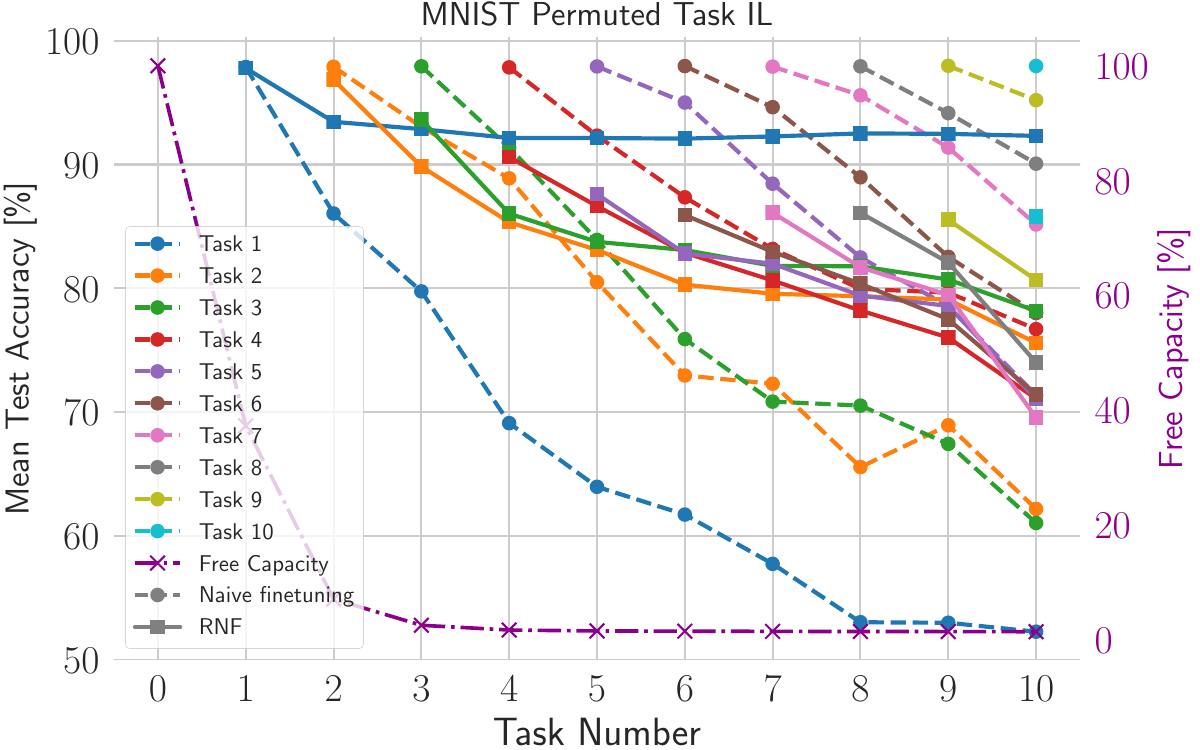}
    \label{fig:MNIST-Permuted:mean_task_acc}}
    \caption{Results on the MNIST dataset. The left \ft{vertical} axis shows the mean test accuracy over all already seen tasks. The \ft{horizontal} axis shows task progression. The right \ft{vertical} axis shows the model’s free capacity.
    \ft{(\protect\subref{fig:MNIST:split_mean_acc}) shows the average test accuracy progression for both \sweb{the naive finetuning} and \sami{\gls{rnf}} on the MNIST Split dataset.}
    (\protect\subref{fig:MNIST:split}) shows the mean test accuracy progression for each task on the MNIST Split dataset.
    (\protect\subref{fig:MNIST-Permuted:avg_acc}) shows the \ft{average} test accuracy progression for both \sweb{the naive finetuning} and \sami{\gls{rnf}} on the MNIST Permuted dataset.
    (\protect\subref{fig:MNIST-Permuted:mean_task_acc}) shows the mean test accuracy progression for each task on the MNIST Permuted dataset.}%
    \label{fig:MNIST}%
\end{figure}

Even though applying \sami{\gls{rnf}} slightly lowers the mean accuracy from 76,99\% to 75\% compared to to the \sami{naive finetuning} baseline, 
Figure \ref{fig:MNIST-Permuted:avg_acc} shows that the average accuracy over all \emph{seen} tasks is above the \sami{naive finetuning} baseline during training. For a closer inspection of task performance, Figure \ref{fig:MNIST-Permuted:mean_task_acc} shows the individual task accuracy over the training. It can be seen that especially task one, two and three benefited the most from \sami{\gls{rnf}}, whereby the small capacity model did not suffice to successfully learn more tasks. Nevertheless, it suggests that parameters can be re-used for new tasks: even though the free capacity drops to below ten percent after the first two tasks, the model can still learn the remaining tasks with reasonable accuracy by an apparent re-utilization of the frozen filters that have been deemed relevant for the previous tasks.

\subsection{CIFAR10 and CIFAR100}

\begin{figure}[ht]
    \centering
    \subfloat[\centering Average Test Accuracy]{\includegraphics[width=0.5\columnwidth]{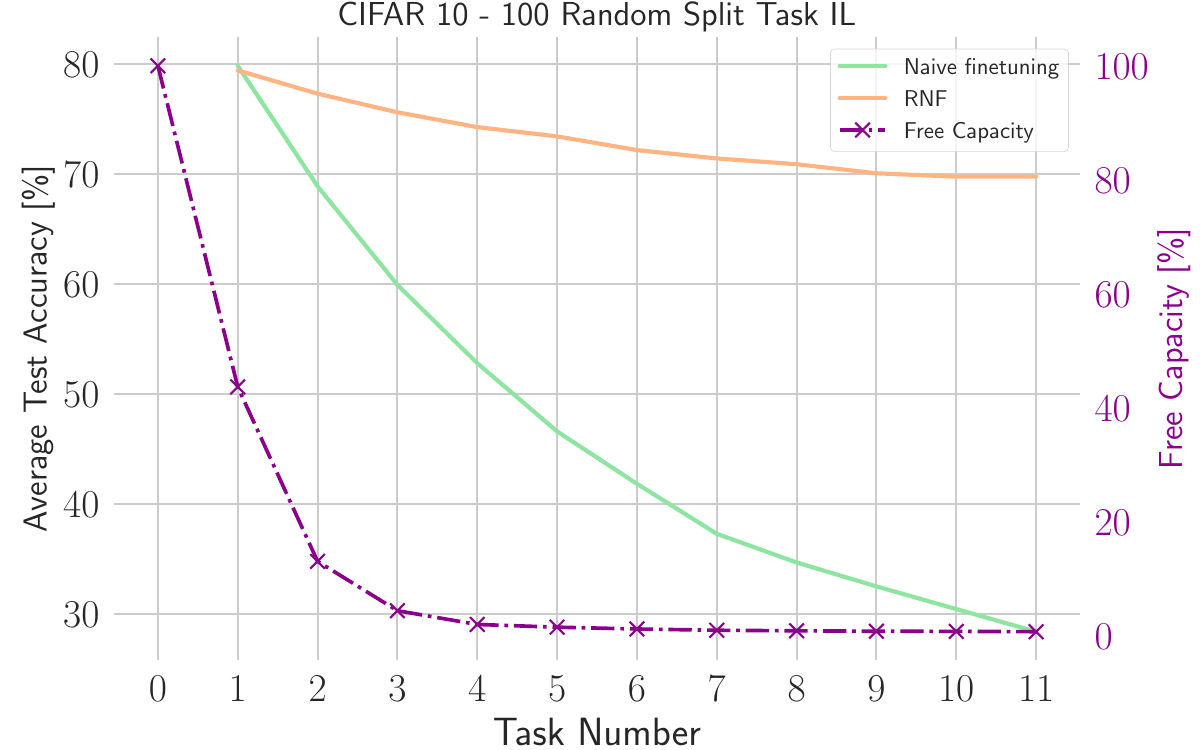}
    \label{fig:CIFAR10-100:random_avg_acc}} %
    \subfloat[\centering Mean test accuracy per task]{\includegraphics[width=0.5\columnwidth]{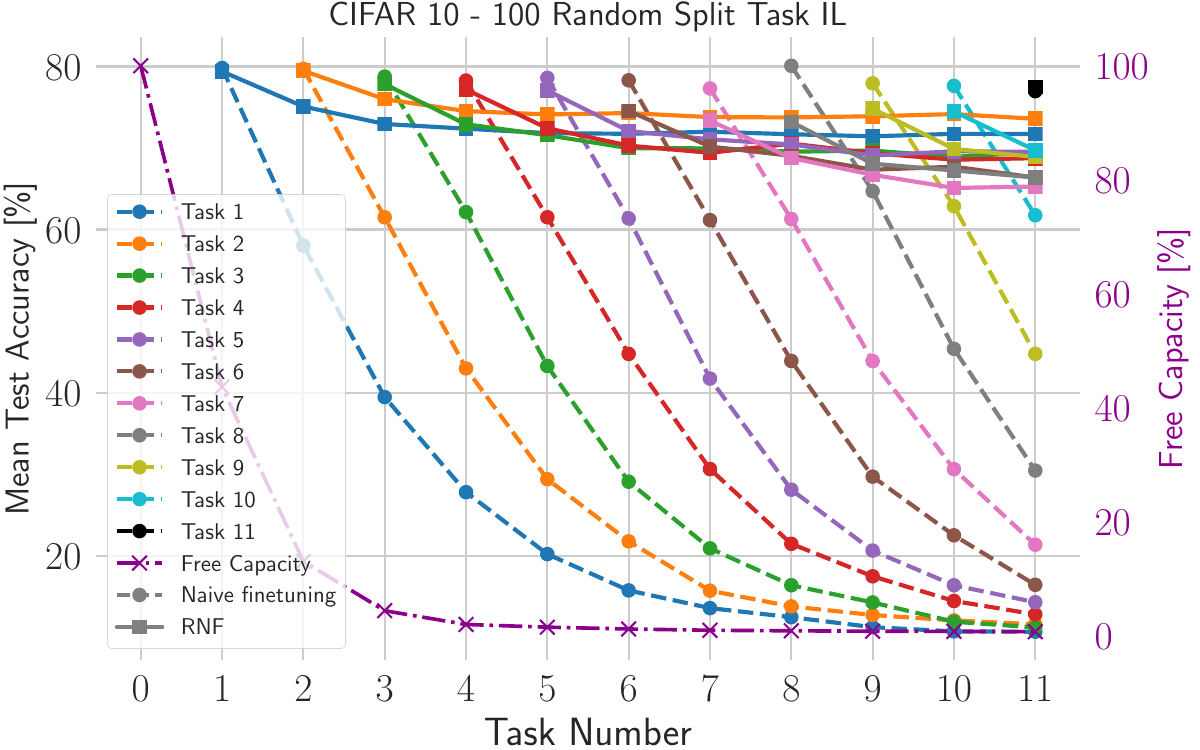}
    \label{fig:CIFAR10-100:random_split}} %
    \\
    \subfloat[\centering Average Test Accuracy]{\includegraphics[width=0.5\columnwidth]{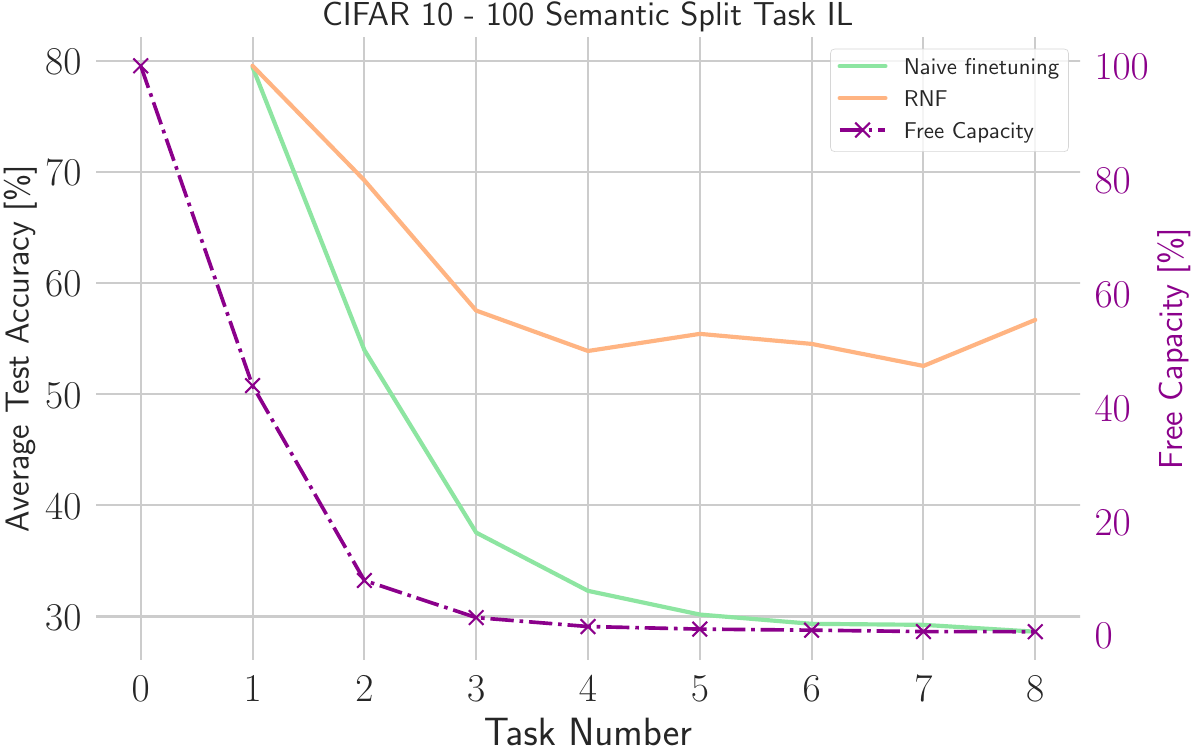}
    \label{fig:CIFAR10-100:avg_acc}} %
    \subfloat[\centering Mean test accuracy per task]{\includegraphics[width=0.5\columnwidth]{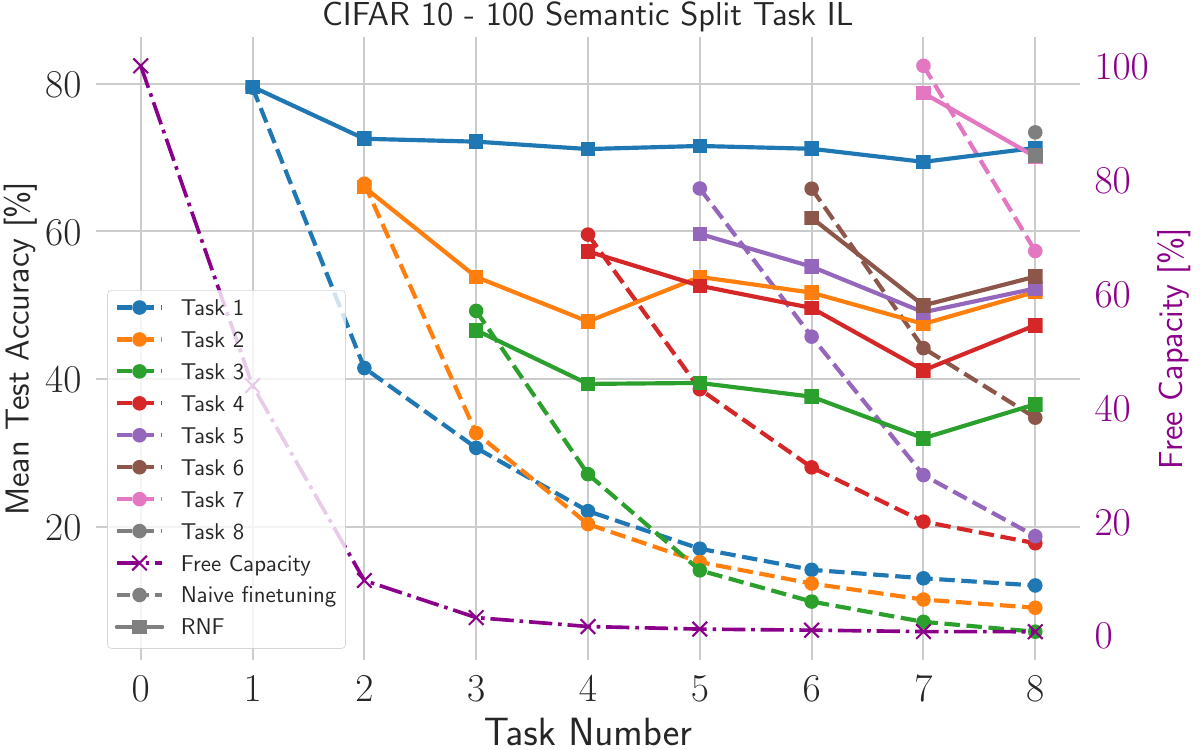}
    \label{fig:CIFAR10-100:mean_task_acc}}%
    \caption{CIFAR100 is split into 10 tasks for the random split and 8 tasks for the semantic split. In the semantic split setup, each task contains a different number of semantically similar classes. The left \ft{vertical} axis shows the mean test accuracy over all already seen tasks. The \ft{horizontal} axis shows task progression. The right \ft{vertical} axis shows the model’s free capacity.
    For \ft{(\protect\subref{fig:CIFAR10-100:random_avg_acc}), CIFAR100 was split randomly into 10 tasks\sweb{,} each containing 10 classes. The plot shows the average test accuracy over all previous tasks after introducing each new task on CIFAR100.
    (\protect\subref{fig:CIFAR10-100:random_split}) shows the mean test accuracy progression for each task on the random split.}
    (\protect\subref{fig:CIFAR10-100:avg_acc}) shows the average test accuracy progression for both \sweb{the naive finetuning} and \sami{\gls{rnf}} on the semantic split of CIFAR100.
    (\protect\subref{fig:CIFAR10-100:mean_task_acc}) shows the mean test accuracy progression for each task on the semantic split.}%
    \label{fig:CIFAR10-100:Semantic}%
\end{figure}

For this dataset, the networks' first task is \slap{to solve the prediction problem defined by} the entire CIFAR10 dataset, after which five further tasks are trained sequentially, each containing 10 randomly selected classes from the CIFAR100 dataset. We expand on this approach and instead split CIFAR100 into ten tasks, each containing ten random classes. Using \sami{\gls{rnf}}, we achieve \ft{an increase of more than 40\% of mean test accuracy compared to the \slap{naive finetuning} baseline as shown in Figure \ref{fig:CIFAR10-100:random_avg_acc}}. 
This effect is also displayed in Figure \ref{fig:CIFAR10-100:random_split}. Even though the network capacity limit seems to be reached at task 5, as in previous experiments, the model's ability to still learn various tasks suggests that the knowledge attained in the previous tasks is enough to facilitate the learning of the remaining classes due to the random assignment of classes to the individual tasks which prevents a semantic bias towards the underlying concepts of the classes within the tasks. Again, the ability of the model to learn new tasks despite a limited residual free capacity signals a high amount of filter re-use of the already frozen filters. 

Another, more complex experiment is performed by \sami{manually} ordering the classes in semantic groups with seven superclasses, containing several subclasses each: Flowers and trees, \slap{(land)} animals, aquatic mammals and fish, random objects, small insects, nature scenes and vehicles. Like in the random grouping experiment, applying \sami{\gls{rnf}} gains almost 30\% in accuracy compared to the \slap{baseline, as \sweb{demonstrated}} in Figure~\ref{fig:CIFAR10-100:avg_acc}. Again, it can be seen that the model's free capacity is getting close to 0\% after training the first four tasks, but the model is able to re-use previous abstractions from previous tasks to still learn the remaining tasks despite the difficulty of the semantic grouping. While the network is not able to achieve an initial accuracy for new tasks that is as high as the \slap{naive finetuning} baseline (as shown in Figure \ref{fig:CIFAR10-100:mean_task_acc}), \sami{\gls{rnf}} can strongly mitigate the loss of accuracy that default finetuning displays when even more tasks are learned. \sweb{In fact, the discrepancy between random grouping (Figure \ref{fig:CIFAR10-100:random_split}) and semantic grouping (Figure \ref{fig:CIFAR10-100:mean_task_acc}) strongly supports the hypothesis of frozen filters being reused for later tasks, since the random grouping has less semantic bias, making filters obtained while learning the first tasks far more useful in subsequent tasks than when grouping semantically.}

\subsection{ImageNet Split}

\begin{figure}[ht]
    \centering
    \subfloat[\centering Task mean test accuracy: ImageNet split]{\includegraphics[width=0.5\columnwidth]{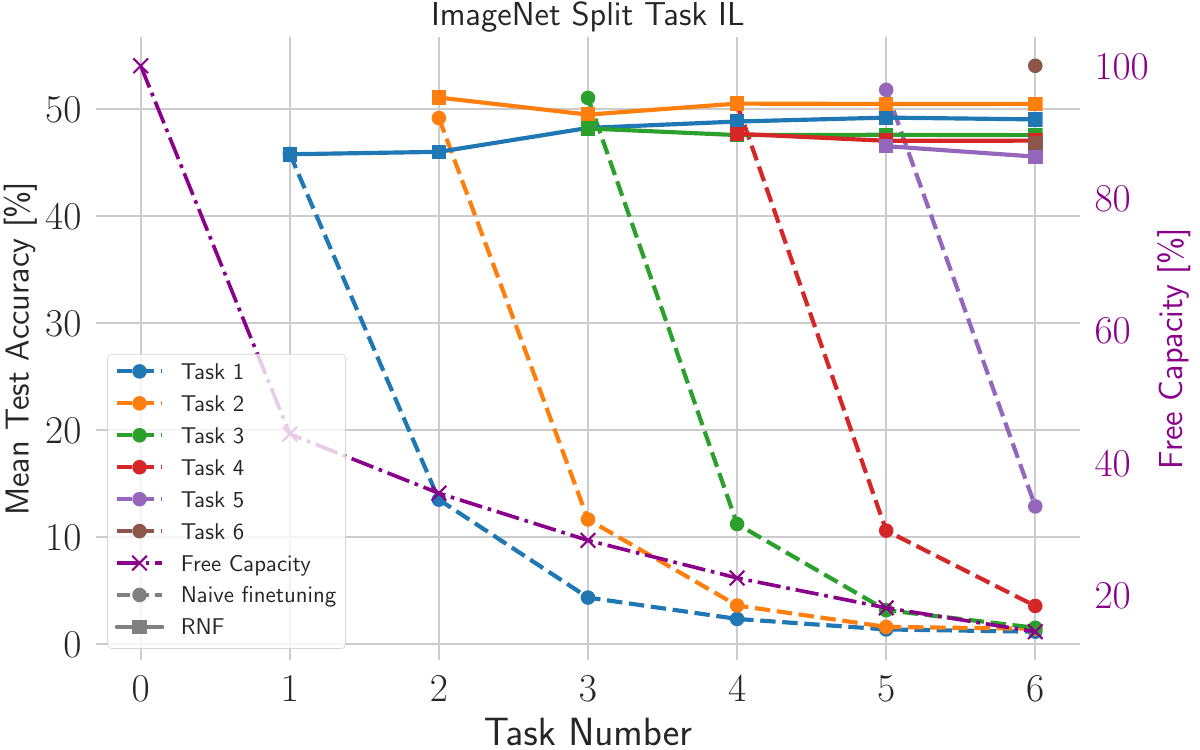}
    \label{fig:ImageNet:1}}%
    \subfloat[\centering Task mean test accuracy: ImageNet-Adience split]{\includegraphics[width=0.5\columnwidth]{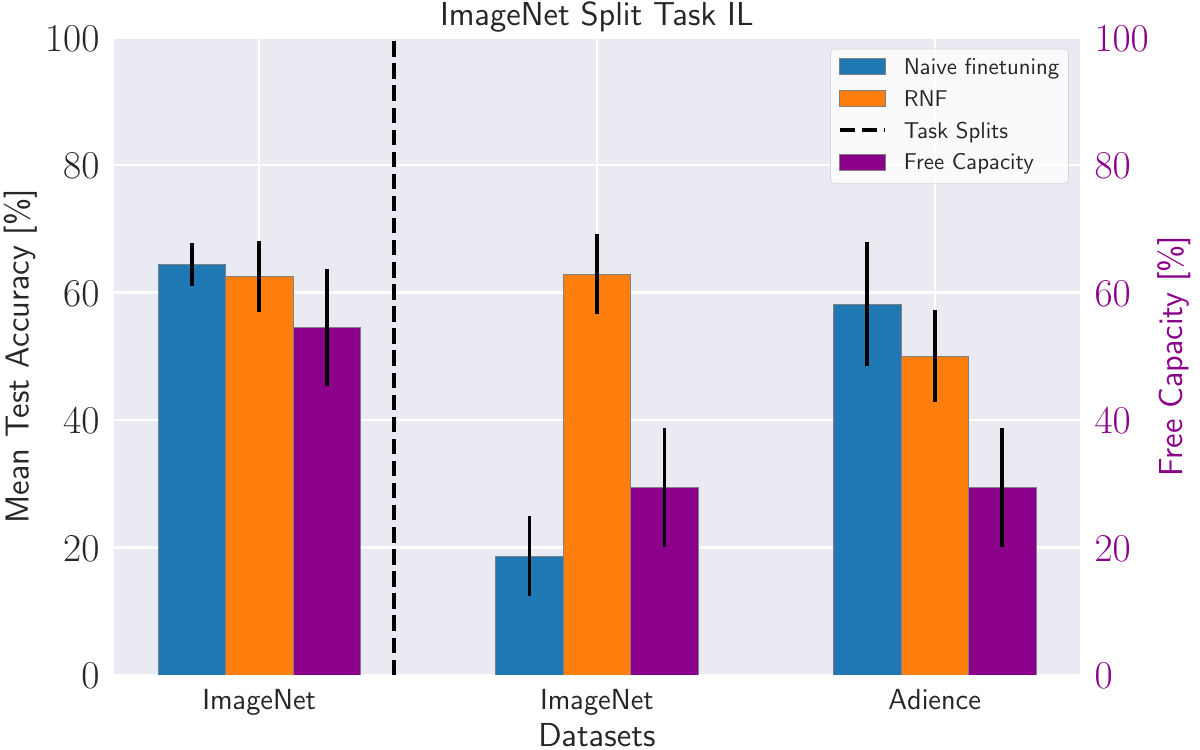}
    \label{fig:AdienceImageNet:1}}
    \caption{(\protect\subref{fig:ImageNet:1}): ImageNet split, where ImageNet is split into 6 sequential tasks consisting of 100 classes each. The left \slap{vertical }axis shows the mean test accuracy over each already seen task. The \slap{horizontal} axis shows the task progression. The dotted lines represent the test accuracy using the default \sami{naive finetuning} approach. The solid lines show the results obtained using \sami{\gls{rnf}}. The right vertical axis shows the model’s free capacity. 
    (\protect\subref{fig:AdienceImageNet:1}): Task-IL protocol by sequentially training both ImageNet and Adience datasets. The left \ft{vertical} axis shows the mean test accuracy of a specific task over five repetitions. The standard deviation is shown as a solid \slap{vertical} black line \slap{at the} top of the bar plot. The \ft{horizontal} axis is split into two groups. The bar plots on the left side of the dotted line represent the mean test accuracy of the model on ImageNet. The bar plots on the right side show the test accuracy of ImageNet after training the Adience dataset and that of the Adience dataset. The test accuracy is computed for both the default baseline approach and \sami{\gls{rnf}}, the model’s free capacity after each task training is shown on the right \ft{vertical} axis.}
    \label{fig:ImageNet}%
\end{figure}

    For the ImageNet \cite{deng2009imagenet} split the \slap{naive finetuning} baseline displays significant catastrophic forgetting even after training each task for only 10 epochs, as is evident in Figure \ref{fig:ImageNet:1}. On this dataset, our method not only preserves the performance of the model on previous tasks but even leads to an increase in accuracy, which in turn results in an overall gain of 35.31\% of accuracy over all tasks \sami{compared to the naive finetuning}. 

\subsection{Adience}
The Adience benchmark dataset of unfiltered faces~\cite{EidingerAdience2014} is a  dataset made up about 26.000 photos of human faces with binary gender- and eight different age group labels. The images are shot under real-world conditions, meaning different variations in pose, lighting conditions and image quality.
We performed experiments on Adience in two scenarios: 
\begin{itemize}
\item \textit{Split}: The dataset is split into two tasks, each consisting of a four-class classification problem of different age groups. \sami{The classes are grouped in a mixed and an ordered setup (Adience-Mixed: Classes [0, 2, 4, 6] and [1, 3, 5, 7], Adience-Ordered: Classes [0-3] and [4-7])}. \sweb{As the classes are labeled with the corresponding age ranges increasing, the groups in the ordered mixing are expected to have more common features.} The baseline is established by finetuning the model for three epochs per sequential task. 
\item \textit{Entire Dataset}: In this scenario, the pretrained model was first pruned to retain the knowledge of ImageNet and then trained on the entire Adience dataset. As Adience is a very complex task, the pruning threshold is increased to 5\% in order to increase the amount of free network capacity in the pretrained model. The model was finetuned on Adience for six epochs.
\end{itemize}

Similar\slap{ly} to the experiments on the benchmark datasets, the results of the proposed method display a lower individual accuracy of the second task compared the \slap{naive finetuning} baseline for both Split tasks, which can be seen in Table \ref{table:adience_ordered_mixed}. Nevertheless, the baseline displays significant catastrophic forgetting, especially for the ordered setup with a drop of almost 46\% for the previously learned task that is almost completely prevented when using \sami{\gls{rnf}}. The lower accuracy scores on task two can be explained by \slap{a} stability\slap{-vs-}plasticity dilemma: decreasing the plasticity of parts of the network can increase stability for already acquired knowledge, but can slow down learning of new tasks, so that with the same amount of training epochs, the task is not learned to the same degree. 
Even though the accuracy of the second task is lower, \sami{\gls{rnf}} still shows that the application of the method is advantageous for sequential tasks regardless of complexity of tasks and size of the model\slap{,} as the mean accuracy over both tasks increases by about 18\% \sami{compared to the naive finetuning}. In both setups, the model retains about 30\% of \slap{free} capacity, making it possible to learn further tasks.

\begin{table}[ht]
    \caption{Average test accuracy on Adience for both the ordered and the mixed setup in two tasks alongside the model's free capacity after each freezing stage. The first ``test accuracy'' column represents the accuracy of the respective task after training the first task. The second ``test accuracy'' column \slap{reports the} test accuracy of the respective task after training the second task.}
    \centering
    \begin{tabular}{|l|l|l|l|l|l|}
        \hline \multicolumn{5}{|c|}{\textbf{Ordered}} \\
          \hline Approach & Current Task & \multicolumn{2}{c|}{Test Accuracy}& Free Capacity  \\
        \hline \hline \slap{Naive Finetuning} & Task 1& $64.38 (\pm 3.4)$& $18.7 (\pm 6.24)$ & $100$\\
        \hline \hline \sweb{Naive Finetuning} & Task 2& $-$ & $58.15 (\pm 9.76)$ & $100$\\
         \hline \noalign{\vskip3\arrayrulewidth} \cline{1-4} \multicolumn{3}{|c|}{\textit{Average over both}} &  $38.43 (\pm 8)$  \\
         \cline{1-4} \noalign{\vskip3\arrayrulewidth}
         \hline \sami{\gls{rnf}} & Task 1 & $62.52 (\pm 5.6)$& $62.87 (\pm 6.3)$& $54.53 (\pm 9.16)$\\
         \hline \hline \sami{\gls{rnf}} & Task 2 & $-$ & $50.02 (\pm 7.23)$& $29.4 (\pm 9.4)$\\
         \hline \noalign{\vskip3\arrayrulewidth} \cline{1-4} \multicolumn{3}{|c|}{\textit{Average accuracy over both tasks}} &  $\textbf{56.45}(\pm \textbf{6.7})$ \\
         \cline{1-4} \noalign{\vskip3\arrayrulewidth} \hline \multicolumn{5}{|c|}{\textbf{Mixed}} \\
        \hline Approach & Current Task & \multicolumn{2}{c|}{Test Accuracy}& Free Capacity  \\
        \hline \hline \sweb{Naive Finetuning} & Task 1&$68.3(\pm 4.33)$ &$52.71(\pm 9.6)$ & $100$\\
        \hline \hline \sweb{Naive Finetuning} & Task 2& $-$ &$80.42(\pm 4.47)$ & $100$\\
        \hline \noalign{\vskip3\arrayrulewidth} \cline{1-4} \multicolumn{3}{|c|}{\textit{Average accuracy over both tasks}} &  $66.57(\pm 7)$  \\
        \cline{1-4} \noalign{\vskip3\arrayrulewidth} \hline \sami{\gls{rnf}} & Task 1 & $70.89(\pm 4.33)$& $70.11(\pm 3.63)$& $47.57(\pm 2.8)$\\
        \hline \hline \sami{\gls{rnf}} & Task 2 & $-$ & $75.85(\pm 4.24)$& $28.8(\pm 3.76)$\\
        \hline \noalign{\vskip3\arrayrulewidth} \cline{1-4} \multicolumn{3}{|c|}{\textit{Average accuracy over both tasks}} &  $\textbf{72.98}(\pm \textbf{3.9})$  \\
        \cline{1-4} \noalign{\vskip3\arrayrulewidth}
    \end{tabular}

    \label{table:adience_ordered_mixed}
\end{table}

As can be seen in Figure \ref{fig:AdienceImageNet:1}, preserving the knowledge of the ImageNet dataset in the pretrained model requires about 73\% of the model's full capacity. The results of \sami{\gls{rnf}} on the Adience-ImageNet split can be found in Table \ref{table:AdienceImageNet}: While the accuracy of task two after our method is again lower than the \slap{naive finetuning} baseline, the accuracy of task one only drops about 2\% compared to almost 20\% in the \slap{baseline} case, granting a mean accuracy increase of 5.6\% over both tasks while still retaining about 15\% free capacity in the model that can be used to learn further tasks.

\begin{table}[!ht]
    \caption{Average test accuracy on splitting ImageNet and Adience into two tasks  alongside the model's free capacity after each pruning stage. The first ``test accuracy'' column represents the test accuracy after training ImageNet. The second column \slap{reports the} test accuracy of ImageNet after training on Adience and the test accuracy on Adience.}
    \centering
    \begin{tabular}{|l|l|l|l|l|l|}
        \hline Approach & Current Task & \multicolumn{2}{c|}{Test Accuracy}& Free Capacity  \\
        \hline \hline \sweb{Naive Finetuning} & ImageNet& $71.5(\pm 0)$&$51.51 (\pm 3.42)$ & $100$\\
        \hline \hline \sweb{Naive Finetuning} & Adience& $-$ &$51.02(\pm 1.66)$ & $100$\\
        \hline \noalign{\vskip3\arrayrulewidth} \cline{1-4} \multicolumn{3}{|c|}{\textit{Average accuracy over both tasks}} &  $51.27(\pm 2.54)$  \\
        \cline{1-4} \noalign{\vskip3\arrayrulewidth} \hline \sami{\gls{rnf}} & ImageNet& $71.5(\pm 0)$& $69.05(\pm 0.16)$& $26.5(\pm 0)$\\
         \hline \hline \sami{\gls{rnf}} & Adience & $-$ & $44.66(\pm 1.61)$& $14.9(\pm 0.02)$\\
         \hline \noalign{\vskip3\arrayrulewidth} \cline{1-4} \multicolumn{3}{|c|}{\textit{Average accuracy over both tasks}} &  $\textbf{56.86}(\pm \textbf{0.89})$  \\
         \cline{1-4} \noalign{\vskip3\arrayrulewidth}
    \end{tabular}

    \label{table:AdienceImageNet}
\end{table}

\subsection{Qualitative Results}
\begin{figure}[!ht]
    \centering
    \subfloat[]{\includegraphics[width=0.4\columnwidth]{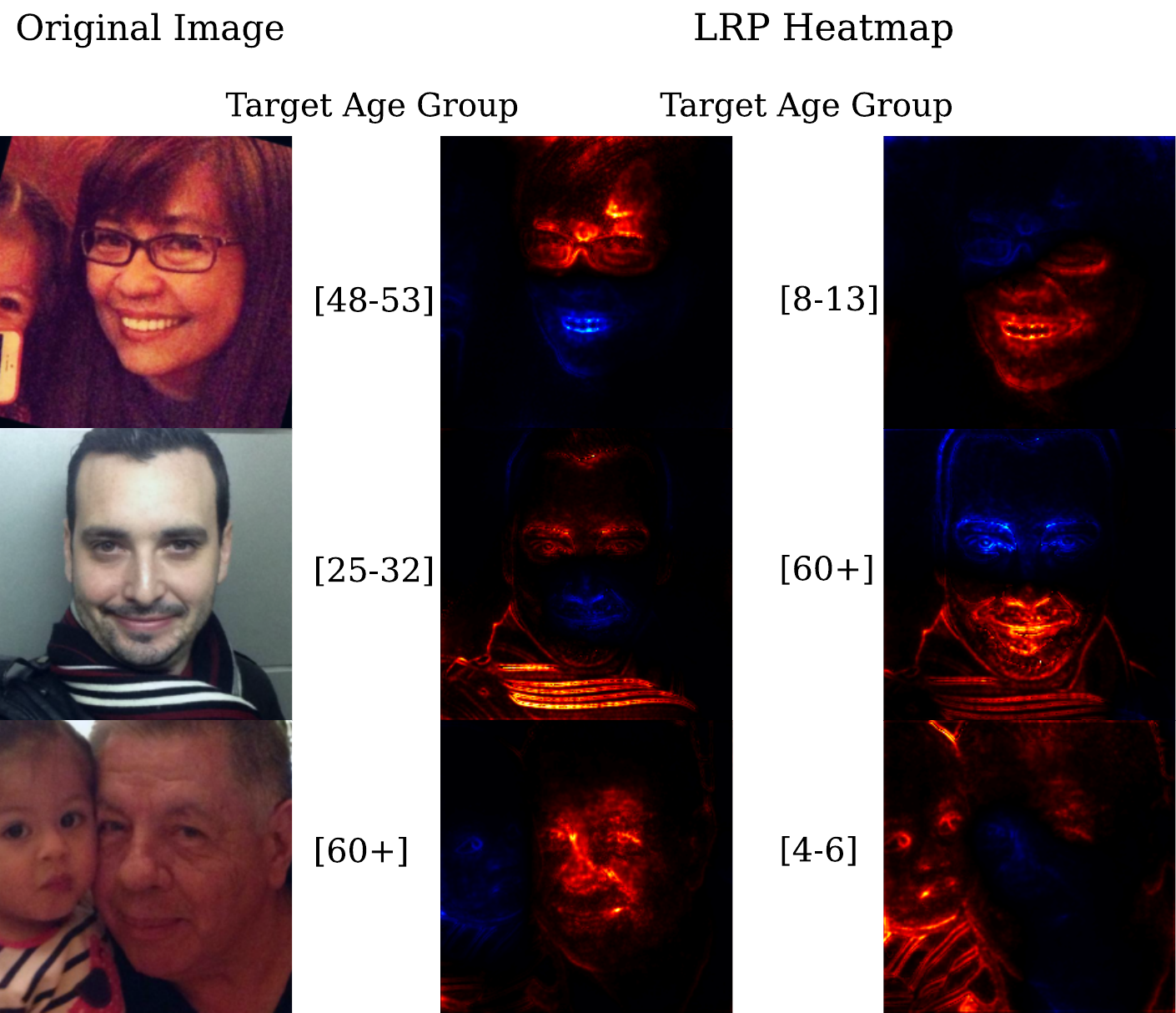}
    \label{fig:Heatmaps:1}}%
    \qquad
    \subfloat[]{\includegraphics[width=0.4\columnwidth]{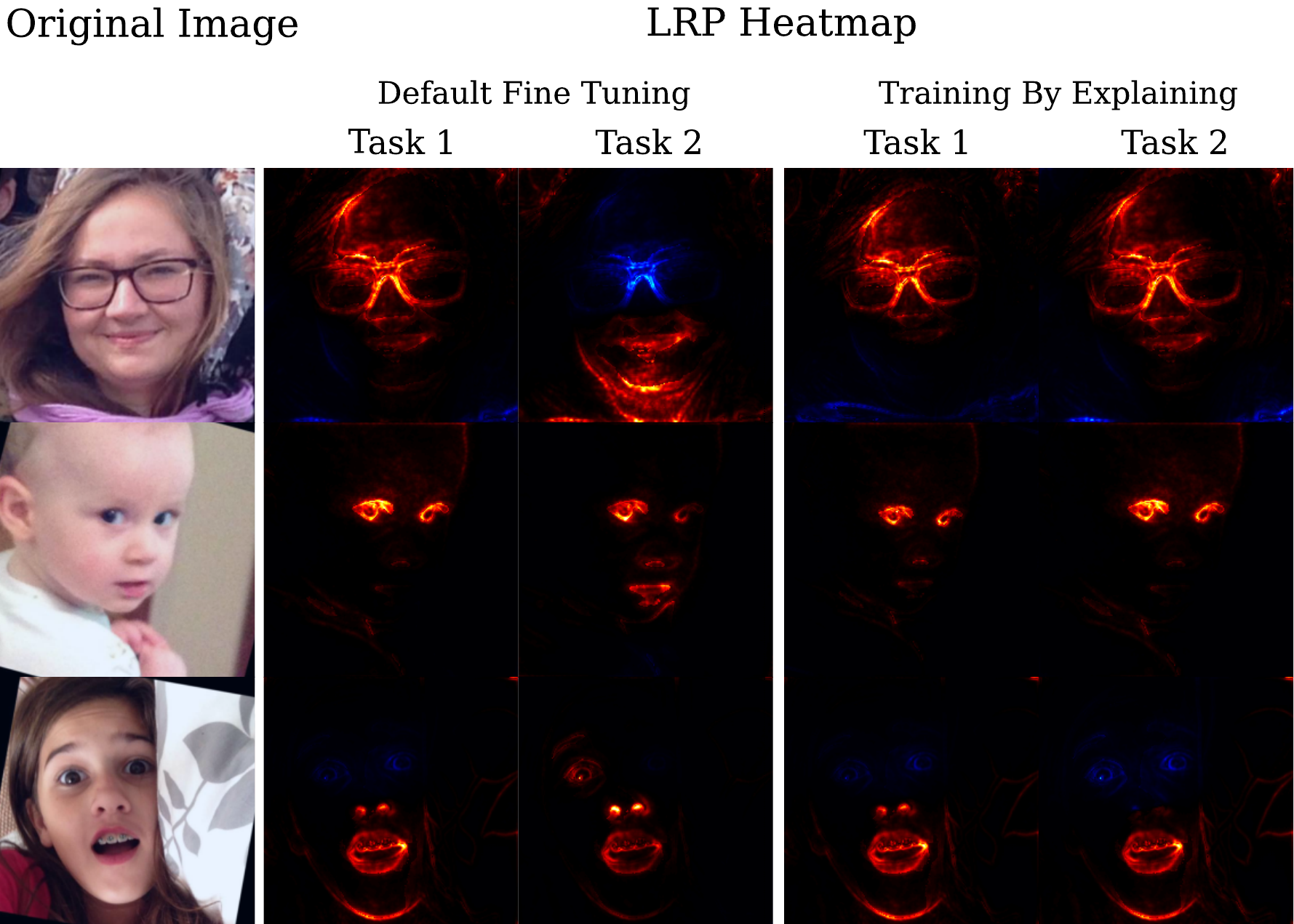}
    \label{fig:Heatmaps:2}} %
     \caption{(\protect\subref{fig:Heatmaps:1}): Images from the Adience dataset alongside their explanations. The \slap{hot} colored regions in the heatmaps \slap{mark} relevant features used by the model for recognizing the chosen class, while \slap{coldly colored regions}   \slap{show} negative relevance, marking contradicting evidence. The relevance is computed \sweb{\wrt} the target class labels indicated on the left of each heatmap. Choosing different target classes produces different explanations, as different class outputs of the model utilize the presented input information differently. 
    (\protect\subref{fig:Heatmaps:2}): Original images from the Adience dataset alongside their explanations for the true class, after either using \slap{naive} finetuning over several tasks\sweb{,} or \gls{rnf}. The figure shows samples from task 1 before and after learning task 2 in the ordered split experiment. This demonstrates that \sami{\gls{rnf}} prevents catastrophic forgetting, \ie, a drift in the reasoning of the model that occurs during \slap{naive} finetuning under continued training for tasks already optimized.}%
    \label{fig:Heatmaps}%
\end{figure}

Alongside the experiments, we also observe changes in the \slap{attribution} heatmaps computed by LRP before and after the application of \sami{\gls{rnf}}. Figure \ref{fig:Heatmaps:1} shows heatmaps computed \wrt different age groups as targets alongside the original image. \slap{Hot pixel color (red over yellow to white)}denote\slap{s} positive relevance towards the target class\sweb{, \slap{\ie, marking features causing the model to predict in favor of the target class, }} while \slap{cold pixel color (blue over turquoise to white)} denote\slap{s} negative relevance. The model learns to associate different facial features with specific age groups, which is especially apparent in the first image, where positive relevance is assigned \slap{consistently} to the glasses for the age group \ft{(48 - 53)}\sweb{,} while negative relevance is assigned in the upper area of the face when the target age group is \ft{(8 - 13)}. The effect of catastrophic forgetting on the relevance distribution can be observed in Figure \ref{fig:Heatmaps:2}. Computing the relevance after training the second task shows that using the \slap{naive finetuning} baseline, the model now assigns negative relevance to the upper area of the face that was previously considered a positive class trait. \sami{\gls{rnf}} retains the initial\slap{ly learned features and prediction strategies of the model, which reflects in the} relevance assignments \slap{which} still focus on the glasses of the woman. Similar behavior is shown in the other images: while the assignment of relevance changes after training with the default baseline, the model that used \sami{\gls{rnf}} is still focusing on the areas that were relevant before training the second task and keeps the sign of the relevance consistent. Finally, the generated heatmaps are consistent with the test accuracy results displayed in the Figures \ref{fig:ImageNet:1} and \ref{fig:AdienceImageNet:1}, and illustrate how previously learned features are preserved despite the introduction of new classes when \sami{\gls{rnf}} is employed.

\section{Conclusion}
Overcoming catastrophic forgetting is one key obstacle towards achieving reliable lifelong learning. Retraining the model from scratch every time new data or tasks are added is sometimes possible, but very inefficient.
In order to prevent the model from forgetting previously learned information, the plasticity of important neurons can be lowered \slap{for further training,} so that they retain the ability to solve earlier tasks. We present an effective algorithm that uses \gls{lrp} to identify the neurons that contribute the most to a given task, \sami{showing a significant increase in accuracy compared to naive finetuning\slap{, while only introducing minimal additional computation cost and requirements on data availablility for previously optimized tasks}}. 
Evaluation of the proposed method on the CIFAR10-100 split achieved an increase in accuracy of about 40\% compared to the baseline, which could also be achieved in a semantic split setting.
The success of our method \sweb{was also} demonstrated on larger datasets, achieving a 35\% increase of accuracy across all sequential tasks when trained on ImageNet. 
In addition to \slap{MNIST, CIFAR and ImageNet}, we \slap{further} evaluated three scenarios on the Adience dataset. We \slap{were able to} show that our method not only \slap{performs favourably} with difficult and unbalanced data\slap{,} but also in a multi-dataset scenario. Retaining the knowledge after training on ImageNet to learn the entire Adience dataset conserved 18\% of accuracy compared to the \slap{naive finetuning} baseline, achieving a net gain of about 11\% \slap{in accuracy} over both tasks. We were able to show that \textit{\sami{\gls{rnf}}} is scalable, efficient\sweb{,} and \slap{effective for rendering neural networks} resilient against catastrophic forgetting in sequential learning setups. Our technique additionally allows for the functional annotation of neural networks. After identifying the relevant parts of the model for a specific task, the obtained learning rate mask could be shared with other researchers, allowing them to employ transfer learning that benefits from existing knowledge through feature reuse, leveraging free network capacity for new tasks while not losing the ability to solve already learned ones. As \gls{lrp} is applicable to a wide range of network architectures, \slap{our RNF} technique can also be applied \slap{beyond the image domain}, \eg\sweb{,} \slap{for} \slap{natural} language processing, \ft{or in Reinforcement Learning. In the latter especially, the impact of non-i.i.d. data is significant, and existing solutions, \eg, experience replay, are highly inefficient. Here, exploring RNF-based solutions is an interesting prospect for future work.}

 \section*{Acknowledgment}
 This work was supported by the German Ministry for Education and Research as BIFOLD (ref.\ 01IS18025A and ref.\ 01IS18037A), the European Union’s Horizon 2020 programme (grant no.\ 965221 and 957059), and the Investitionsbank Berlin under contract No.\ 10174498 (Pro FIT programme).

\bibliographystyle{splncs04}
\bibliography{bibliography.bib}

\setcounter{figure}{0}
\setcounter{table}{0}
\setcounter{page}{1}
\pagenumbering{roman}
\renewcommand{\figurename}{Supplementary Figure}
\renewcommand{\tablename}{Supplementary Table}
\renewcommand\thefigure{\arabic{figure}}
\renewcommand\thetable{\arabic{table}}

\begin{appendix}

\section{Appendix}
\label{appendix:expdet}

\subsection{MNIST-Split}
\label{appendix:expdet:mnistsplit}
The model architecture was taken from \cite{DBLP:journals/corr/abs-1904-07734}, which compares multiple methods for mitigating catastrophic forgetting. It consists of two hidden layers with 400 neurons each and ReLU activations. As for most experiments except the real-world dataset, the pruning threshold is set to 2\%, meaning that the accuracy can drop by up to 2\% before the pruning procedure is halted. We use the Adam optimizer with a learning rate of 0.001, $\beta_1 = 0.9$ and $\beta_2=0.999$ with a batch size of 128.

\subsection{MNIST-Permuted}
\label{appendix:expdet:mnistperm}
For this experiment, the architecture from \cite{samek2017understanding} was adapted by increasing the number of hidden layer units to 1000 per layer to match the increased complexity of the task. Additionally, the learning rate was decreased to $0.0001$ and the model was trained for ten instead of four epochs per task. 

\subsection{CIFAR10 and CIFAR100}
\label{appendix:expdet:CIFAR}
In this experiment, we adopted architecture and experimental setup from \cite{Zenke2017synapticintelligence}.

\subsection{ImageNet Split}
Here, we replicate the conditions from \cite{wu2019incremental} but establish our baseline after ten instead of 70 epochs, which we also use when \sweb{applying \gls{rnf}}.

\subsection{Adience}
As is state-of-the art for this dataset \cite{EidingerAdience2014}, we normalize to zero mean and unit standard deviation during training and apply data augmentation \sweb{for the training data }by randomly cropping to 244x244 as well as horizontal flipping. For \sweb{testing}, each sample is cropped five times to 244x244 \sweb{(four corner crops and one center crop)}, where each crop is additionally mirrored \sweb{horizontally}. The ground truth is then compared to the mean of the Softmax activations of the ten samples. \sweb{Only the center crops are used in the reference data.}  As the dataset is strongly imbalanced, we additionally employ a resampling strategy during training that undersamples classes with a high number of samples and oversamples classes with a low number of samples by computing the class probabilities and then sampling from a multinomial distribution.

In this experiment, we employ a VGG-16 network architecture \cite{simonyan2015convolutional} that has been pretrained on ImageNet (from the PyTorch \cite{paszke2019pytorch} model zoo), as well as an Adam optimizer and L2 regularization.

\begin{itemize}
\item \textit{Split}: We used a learning rate of $0.0001$, L2 regularization with $\lambda = 0.01$ and a batch size of 32.\\
\item \textit{Entire Dataset}: The model was trained with a learning rate of $0.00001$ and $\lambda = 0.001$.
\end{itemize}

\end{appendix}

\end{document}